# Persian Sentiment Analyzer: A Framework based on a Novel Feature Selection Method

**Ayoub Bagheri**[1] and **Mohamad Saraee**[2]

[1] Intelligent Database, Data Mining and Bioinformatics Lab,
Computer Engineering Department, Isfahan University of Technology, Isfahan, Iran
Email: a.bagheri@ec.iut.ac.ir

[2]School of Computing, Science and Engineering, University of Salford, Manchester, UK
Email: m.saraee@salford.ac.uk

**ABSTRACT**

*In the recent decade, with the enormous growth of digital content in internet and databases, sentiment analysis has received more and more attention between information retrieval and natural language processing researchers. Sentiment analysis aims to use automated tools to detect subjective information from reviews. One of the main challenges in sentiment analysis is feature selection. Feature selection is widely used as the first stage of analysis and classification tasks to reduce the dimension of problem, and improve speed by the elimination of irrelevant and redundant features. Up to now as there are few researches conducted on feature selection in sentiment analysis, there are very rare works for Persian sentiment analysis. This paper considers the problem of sentiment classification using different feature selection methods for online customer reviews in Persian language. Three of the challenges of Persian text are using of a wide variety of declensional suffixes, different word spacing and many informal or colloquial words. In this paper we study these challenges by proposing a model for sentiment classification of Persian review documents. The proposed model is based on lemmatization and feature selection and is employed Naive Bayes algorithm for classification. We evaluate the performance of the model on a manually gathered collection of cellphone reviews, where the results show the effectiveness of the proposed approaches.*

## 1. INTRODUCTION

In the recent decade, with the enormous growth of digital content in internet and databases, sentiment analysis has received more and more attention between information retrieval and natural language processing researchers (Bagheri, Saraee, and de Jong, 2013a) (Hu and Liu, 2004a) (Hu and Liu, 2004b) (Liu and Zhang, 2012). Up to now studies in sentiment analysis have covered a wide range of tasks, including sentiment classification on word, phrase, sentence and document level (Cui, Mittal, and Datar, 2006) (Gamon, 2004) (Moraes, Valiati, and Gavião Neto, 2012) (Pang, Lee, and Vaithyanathan, 2002) (Saraee and Bagheri, 2013) (Yussupova, Bogdanova, and Boyko, 2012) (Zhu, Wang, and Mao, 2012), opinion target identification or aspect detection (Bagheri, Saraee, and de Jong, 2013a) (Bagheri, Saraee, and de Jong, 2013b) (Hu and Liu, 2004a) (Liu and Zhang, 2012) (Popescu and Etzioni, 2005) (Qiu, Liu, Bu et al. 2011) (Zhu, Wang, Zhu et al. 2011), opinion word detection and opinion orientation identification (Hogenboom, Boon, and Frasincar, 2012) (Hu and Liu,

2004b) (Liu, Hu, Cheng, 2005) (Liu and Zhang, 2012) (Turney, Littman, 2002). The ability of sentiment classification on multiple levels is important since different applications have different needs. For example an opinion summarization system on product reviews may needs both classifications on document and word level. In this work we focus on classifying product reviews on the document level.

Up to now, many researches have been conducted on English, Chinese or Russian document sentiment analysis or classification (Liu and Zhang, 2012) (Qiu, Liu, Bu et al. 2011) (Yussupova, Bogdanova, and Boyko, 2012). However on Persian text, in our knowledge there is little investigation conducted on sentiment analysis (Saraee and Bagheri, 2013) (Shams, Shakery, and Faili, 2012). Persian is an Indo-European language, spoken and written primarily in Iran, Afghanistan, and a part of Tajikistan. The amount of information in Persian language on the internet has increased in different forms. As the style of writing in Persian language is not firmly defined on the web, there are too many web pages in Persian with completely different writing styles for the same words. Different writing styles produce many challenges about Persian text processing such as, informal and colloquial words, declensional suffixes, multiple types of writing for a word and word spacing (Farhoodi and Yari, 2010) (Saraee and Bagheri, 2013) (Shams, Shakery, and Faili, 2012) (Taghva, Beckley, and Sadeh, 2005). Therefore in this paper, we study the main challenges of Persian language, and experiment our model on a Persian product review dataset.

Feature selection is a process that can be applied in many machine learning applications like optimization problems, search algorithms, and classification and clustering problems (Ali, Alkhatib, Tashtoush, 2013) (Bagheri, Akbarzadeh - T, Saraee, 2008) (Duric, Song, 2012) (Ghani, Probst, Liu et al., 2006) (Precup, David, Petriu et al., 2012). Feature selection is one of the most important parts in a classification problem which removes irrelevant features to decrease the computational cost. Feature selection process works by ranking all the features and then selecting a subset containing best features (Ghani, Probst, Liu et al., 2006) (Mitchell, 1997) (Pei, Shi, Marchese et al. 2007) (Rennie, 2004) (Tang, Shepherd, Milios et al., 2005). Feature selection for sentiment analysis is a very tough optimization work when the feature space of sentiment words is too large. In this work we study three different feature selection methods and we present a new feature selection approach for the sentiment analysis model to improve the overall accuracy. The proposed approach for feature selection is not specifically for Persian text and it is applicable for other languages or domains.

In the reminder of this paper, existing works on sentiment classification will be given in section 2. Section 3 describes the proposed model for sentiment classification of Persian reviews, including the Naive Bayes classifier, different feature selection methods and the proposed approach for feature selection. Subsequently we describe our empirical evaluation and discuss important experimental results in section 4. Finally we conclude with a summary and some future research directions in section 5.

## 2. RELATED WORK

Sentiment classification aims to distinguish whether people like or dislike a product, a service, an organization, an individual or a topic. Sentiment classification for product reviews has recently attracted much attention from the natural language processing community. Researchers have investigated the effectiveness of different machine learning algorithms using various features to develop sentiment classifiers on languages like English or Chinese (Liu and Zhang, 2012) (Qiu, Liu, Bu et al. 2011). On Persian text, there is very little investigation conducted on sentiment analysis. Besides our study (Saraee and Bagheri, 2013) through our energies on searching the web, only one people's work can be found, i.e., Shams et al. (Shams, Shakery, and Faili, 2012).

Shams et al. (Shams, Shakery, and Faili, 2012) employed a new lexical resource for Persian sentiment analysis, PersianClues, with an unsupervised LDA-based sentiment analysis method, LDASA, to categorize each document into its related polarity using a classification algorithm. In their work, to generate the PersianClues, an automatic translation approach is used to translate the existing English clues to Persian. They experiment their model on Persian documents about hotels, cellphones and digital cameras and reached to 76% accuracy on average.

Turney and Littman (Turney, Littman, 2002) employed a vocabulary of adjectives and adverbs to find out about sentiment orientation of each word in the documents. They used point-wise mutual information and latent semantic analysis to calculate the orientation of the extracted words according to their co-occurrences with the seed words, such as *excellent* and *poor*. They determined the polarity of a document by averaging the sentiment orientation of words in the document.

Instead of conducting the analysis on the word level, another stream of research performs sentiment classification on the document or review level. Pang et al. (Pang, Lee, and Vaithyanathan, 2002) conducted an extensive experiment on English movie reviews on the document level using three supervised machine learning methods, Naive Bayes, Maximum Entropy Model and Support Vector Machines. Their results show that machine learning techniques definitely outperform human-produced baselines. Additionally they found that machine learning approaches could not perform as well on sentiment classification as on traditional text classification problem.

Up to now different approaches have been used to sentiment classification problem. Little emphasis has been given to feature selection techniques in sentiment analysis. Gamon (Gamon, 2004) and Yi et al. (Yi, Nasukawa, Bunescu et al., 2003) used log likelihood to select important attributes from a large initial feature space. Duric and Song (Duric and Song, 2012) approached the task of feature selection for sentiment analysis by using a Content and Syntax model, HMM-LDA, to separate the entities in a review document from the potentially sentiment carrying modifiers. Abbasi et al. (Abbasi, Chen, and Salem, 2008) developed an entropy weighted genetic algorithm (EWGA) for efficient feature selection in order to improve accuracy and identify key features for each sentiment class. In their work EWGA significantly outperformed the no feature selection baseline and GA on all their experiments. Hence, Using feature selection could improve accuracy and focus in on a key feature subset of sentiment discriminators.

In this paper, we propose a sentiment classification model for the document level of cellphone reviews with various feature selection methods. In the model we present a new feature selection approach based on the Mutual Information method. We use stemming and n-gram features to overcome obstacles in Persian text.

## 3. PROPOSED MODEL FOR SENTIMENT ANALYSIS

In the proposed sentiment analysis model, after preprocessing, a stemming step is used to remove the redundancy from feature space, then we utilize a feature selection method to obtain better performance with the features and reduce the computational cost and finally we apply a classifier learning algorithm as training step. In order to describe the proposed model for sentiment analysis, in this section, we will discuss about the stemming and learning steps and will introduce the proposed feature selection approach among presenting other feature selection approaches.

### 3.1. Stemming

Stemming or lemmatization is to reduce a word to a more general form, possibly its root. For example, stemming the term "*interesting*" may produce the term "*interest*". Persian language has a complicated morphology. In Persian language suffixes and prefixes are concatenated to words to modify the meaning. Like English nouns, Persian nouns are affixed to signify possession and plurality. On the other hand, Persian verbs are modified more extensively than English verbs. Persian verbs vary according to tense, person, negation, and mood. Therefore, a given verb may have different forms and variations. In this study we re-implemented Taghva Persian stemmer based in the algorithm presented in (Taghva, Beckley, and Sadeh, 2005).

### 3.2. Description of sentiment classifier

In this paper, we consider Naive Bayes algorithm which is a machine learning approach as the sentiment classifier. Mathematically, the problem of classifying review documents based on their sentiments can be represented as follows. There are two classes, the class of positive reviews, $c_1$, and the class of negative reviews, $c_2$:

$$C = \{c_1, c_2\}$$

There is a set of reviews:

$$R = \{r_1, r_2, \ldots, r_n\}$$

And an unknown classification function:

$$F : C \times R \rightarrow \{0,1\}$$

We need to build a classifier $F'$ as close to the classification function F as possible. In this problem we use vector model to represent the feature space. For the feature space we extract n-gram features to deal with the conflicting problem of space and pseudo-space in Persian sentences. Here we use unigram and bigram phrases as n-gram features. Therefore in this model, the sequence of the words is important. For example in the review sentence of a cellphone, "کیفیت آنتن دهی این گوشی خوب است." / The signal quality of this phone is good." in addition to considering each word as a feature individually, we extract bi-gram features: "آنتن دهی / signal", "کیفیت آنتن / signal quality" and "گوشی خوب / good phone".

Experiments show using n-gram features could solve the problem of different word spacing in Persian text.

*Naive Bayes Algorithm*

We used the MAP (Maximum a Posteriori) Naive Bayes algorithm in our experiments as a classifier for Persian sentiment classification system, because Naive Bayes is a kind of important classification algorithm and it has a high speed and is easy to implement rather than approaches like SVM or neural networks (Mitchell, 1997) (Rennie, 2004). In a classification problem, training and test dataset have to be labeled by a human expert and the classifier predicts the class of each data in test dataset. Naive Bayes algorithm assigns a new review document with a class with the maximum probability. This maximum value can be calculated by equation (1):

$$Naive\ Bayes\ Classifier : c_{NB} = argmax_{c_j \in C} P(c_j) \prod_i P(a_i | c_j) \quad (1)$$

Where $c_{NB}$ is the assigned class or output of Naive Bayes algorithm, $c_j$ shows the class j$^{th}$, $P(c_j)$ is prior probability of class j in the set of all classes C and $P(a_i|c_j)$ shows conditional probability of feature *i* in class *j*. Output of the Naive Bayes algorithm is the maximum probability between classes. To calculate $c_{NB}$, we require estimates for the probability terms $P(c_j)$ and $P(a_i|c_j)$. $P(c_j)$ can simply estimate based on the fraction of each class in the training data by equation (2):

$$P(c_j) = \frac{|R_j|}{|R|} \quad (2)$$

Where $|R|$ is total number of review documents and $|R_j|$ is the number of review documents with class label j. For estimating $P(a_i|c_j)$ we define an expression named *FSpace* which is the set of all distinct features in any review document in the dataset. Therefore $P(a_i|c_j)$ can be computed by equation (3):

$$P(a_i|c_j) = \frac{n_{ij} + 1}{n_j + |FSpace|} \quad (3)$$

Where $n_j$ is the total number of features in all training data with the class label of $c_j$, $n_{ij}$ is the number of times feature $a_i$ is found among these n features, and $|FSpace|$ is the size of feature space.

When we focus on the equation of $P(c_j)$ we found that the fraction maybe zero in some circumstances, therefore we use a modified version of the equation (2) in equation (4):

$$P(c_j) = \frac{1 + |R_j|}{|R| + |FSpace|} \quad (4)$$

After training the classifier, we can use the algorithm for estimating class label of a new review document from equation (1).

### 3.3. Feature selection for sentiment analysis

Feature Selection methods sort features on the basis of a numerical measure computed from the documents in the dataset collection, and select a subset of the features by thresholding that measure. In this paper four different information measures were implemented and tested for feature selection problem in sentiment analysis. The measures are Document Frequency (DF), Term Frequency Variance (TFV), Mutual Information (MI) (Yang and Pedersen, 1997) and Modified Mutual Information (MMI).

*Document Frequency*

Document Frequency is the number of documents in the training dataset in which a term or feature occurs. Only the features that occur in a large number of documents are retained. DF thresholding is the simplest technique for feature selection and reduction.

*Term Frequency Variance*

The basic idea of TFV is to rank the quality of a feature based on the variance of its frequency (Tang, Shepherd, Milios et al., 2005). TFV can be calculated by:

$$TFV(f) = \sum_{i=1}^{k}[tf(f,c_i) - mean\_tf(f)]^2 \qquad (5)$$

Where $f$ is current feature, $k$ is number of classes, $c_i$ is $i$th class and $tf(f,c_i)$ is frequency of feature $f$ in all documents of class $c_i$. The intuition of this method is to select features which have less frequency and are common across all classes.

*Mutual information and modified mutual information–the proposed feature selection approach*

In this paper we introduce a new approach for feature selection, Modified Mutual Information. In order to explain Modified Mutual Information (MMI) measure, it is helpful to first introduce Mutual Information (MI) by defining a contingency table (see Table 1).

**Table 1.** Contingency table for features and classes

|   | $C$ | $\bar{c}$ |
|---|---|---|
| $f$ | A | B |
| $\bar{f}$ | C | D |

Table 1 records co-occurrence statistics for features and classes. Therefore we see that the number of times a class $c$ occurred with the presence of feature $f$ in the training dataset is A, for example. We also have that the number of review documents, N = A+B +C +D. These statistics are very useful for estimating probability values (Tang, Shepherd, Milios et al. 2005) (Pei, Shi, Marchese et al. 2007) (Ghani, Probst, Liu et al. 2006).

By using Table 4, MI can be computed by equation (7):

$$MI(f,c) = \log \frac{P(f,c)}{P(f)P(c)} \tag{6}$$

Where $P(f,c)$ is the probability of co-occurrence of feature $f$ and class $c$ together, and $P(f)$ and $P(c)$ are the probability of co-occurrence of feature $f$ and class $c$ in the review documents respectively. Therefore by Table 1, MI can be approximated by Equation (8):

$$MI(f,c) = \log \frac{A*N}{(A+B)*(A+C)} \tag{7}$$

Or for simplicity:

$$MI(f,c) = \frac{A*N}{(A+B)*(A+C)} \tag{8}$$

Intuitively MI measures if the co-occurrence of $f$ and $c$ is more likely than their independent occurrences, but it doesn't measure the co-occurrence of $f$ and $\bar{c}$ or the co-occurrence of other features and class $c$. We introduce a Modified version of Mutual Information as MMI which consider all possible combinations of co-occurrences of a feature and class label. First we define four parameters as the following:

- $p(f,c)$: *Probability of co-occurrence of feature f and class c together.*
- $p(\bar{f},\bar{c})$: *Probability of co-occurrence of all features except f in all classes except c together.*
- $p(\bar{f},c)$: *Probability of co-occurrence of all features except feature f in class c.*
- $p(f,\bar{c})$: *Probability of co-occurrence of feature f in all classes except c.*

We name the first two parameters as positive factors, which by definition they have a positive impact on the finding the relative information between feature $f$ and class $c$. likewise we name the second two parameters as negative factors. Therefore we calculate MMI score as Equation (10):

$$MMI(f,c) = \log \frac{p(f,c)*p(\bar{f}\,\bar{c})}{p(f)*p(c)*p(\bar{f})*p(\bar{c})} - \log \frac{p(\bar{f},c)*p(f,\bar{c})}{p(f)*p(c)*p(\bar{f})*p(\bar{c})} \tag{9}$$

Where $P(f)$ and $P(c)$ are the probability of independent occurrence of feature $f$ and class $c$ in the review documents respectively. $p(\bar{f})$ is the number of review documents which not contain feature $f$ and $p(\bar{c})$ is the number of documents with the classes other than class $c$. MMI considers positive and negative factors to calculate the score between $f$ and $c$. In this measure, the positive factor assigns with a positive coefficient and the negative factor assigns with a negative coefficient, hence the measure uses the information of both positive and negative factors.

Based on Table 1, MMI can be approximated by Equation (11):

$$MMI(f,c) = \log \frac{A*D}{(A+C)*(B+D)*(A+B)*(C+D)} - \log \frac{C*B}{(A+C)*(B+D)*(A+B)*(C+D)} \quad (10)$$

For simplicity, we calculate MMI as:

$$MMI(f,c) = \frac{A*D - C*B}{(A+C)*(B+D)*(A+B)*(C+D)} \quad (11)$$

## 4. EXPERIMENTAL RESULTS

In this Section, we describe the evaluation of the proposed sentiment analysis model with a comparative study on the proposed MMI feature selection approach in a variety of settings in compare to the results of other feature selection approaches. In this paper we chose the Persian as validation language. Hence, in the following, first the challenges in Persian sentiment analysis will be discussed and then data collection, evaluation metrics, use of cross validation and important experimental results will be discussed.

### 4.1. Persian language

Persian text mining or specifically Persian sentiment analysis suffers from low quality. One of the main challenges is the lack of comprehensive solutions or tools for the tasks like stemming, POS tagging, and feature selection. One of the consequences is reaching to a very large vector space, hence the process is very time consuming and leads to weak results. Using of a wide variety of declensional suffixes is another challenge of Persian language, Tables 2 and 3 show examples of different suffixes in Persian text. Another common problem of Persian text is word spacing. In Persian in addition to white space as inter-words space, an intra-word space called pseudo-space separates word's part. Using white space or do not using any space instead of pseudo-space is a great challenge in Persian reviews. For example in the sentence "این گوشی قابلیت تشخیص دست‌خط خوبی دارد." / This phone has good handwritten recognition ability", the word "دست‌خط / handwritten" is a word which uses pseudo-space and contains two other words "دست / hand" and "خط / written line". In this sentence if the algorithm interprets the word "دست‌خط / handwritten" united or separated, the feature space and the results could be different. Another important challenge in customer reviews in Persian language is that of utilizing many informal or colloquial words in text. Table 4 shows some examples for Informal or colloquial words in Persian documents.

**Table 2.** Examples for Present Tense Suffixes in Persian

| Present tense in Persian | English Translation | Suffix |
|---|---|---|
| من می بینم | I see | م |
| تو می بینی | you see | ی |
| او می بیند | he sees | د |
| ما می بینیم | we see | یم |
| شما می بینید | you see | ید |
| آنها می بینند | they see | ند |

**Table 3.** Examples for Plural Suffixes in Persian

| Persian Plural word | English Translation | Suffix |
|---|---|---|
| تصویرها | Pictures | ها |
| دوستان | Friends | ان |
| نکات | Tips | ات |
| مراجعین | Visitors | ین |

**Table 4.** Examples for Informal or colloquial words in Persian

| Informal or colloquial word | Formal form | English Translation |
|---|---|---|
| ازش | از آن | It |
| خونه | خانه | Home |
| گوشیرو | گوشی را | this phone |
| فک کردن | فکر کردن | Think |
| نداره | ندارد | have not |
| باهاش | با آن | by that |

We believe that our model with using n-gram features, stemming and feature selection overcomes to the Persian language challenges and can enormously decrease the size of the vector space and affect the Persian sentiment classification process positively.

**4.2. Data**

To test our methods we compiled a dataset of 1020 online customer reviews in Persian language from different brands of cell phone products including *Nokia, Apple, Samsung, Sony, LG, Motorola, Huawei* and *HTC* (Saraee and Bagheri, 2013). We assigned two annotators to label customer reviews by selecting a positive or negative polarity on the review level. After annotation, the dataset reached to 511 positive and 509 negative reviews.

**4.3. Cross validation**

A supervised learning algorithm needs some of data to be labeled as training data and some of them as test data (Hu and Liu, 2004b). These two datasets must be separate to prevent from false results in evaluating the performances of the methods. Therefore multiple runs of the experiments are usually needed with different datasets at each turn. One of the approaches splitting the dataset and running the experiment is N-fold cross validation. N-fold cross validation consists in splitting the dataset into N subsets of equal size. At each turn, one set is used for testing and the rest for training the system. In our case, 5-fold cross validation is used. At each turn, 4 folds will be used for training and learning and one for testing, in such way that every subset will be used once for testing purposes. Then, the average over all 5 experiments will be as an estimate of the performance of the classifier.

## 4.4. Evaluation metrics

We use precision, recall and F-score to measure the effectiveness of a feature selection method and a sentiment classifier.

**Table 1.** Contingency table for evaluation metrics

|  | Not Predicted as Feature | Predicted as Feature |
|---|---|---|
| Wrong Features | TN | FP |
| True Features | FN | TP |

The precision, recall and F-score are calculated based on Table 5 as:

$$Precision = \frac{TP}{TP+FP} \tag{12}$$

$$Recall = \frac{TP}{TP+FN} \tag{13}$$

$$F-score = \frac{2*(Precision*Recall)}{Precision+Recall} \tag{14}$$

To deal with multiple datasets (products), we adopt the macro and micro average (Yang, 1999) to assess the overall performance. The macro- and micro-averaged precision and recall across the $n$ datasets are defined as follows:

$$Macro-averaged\ precision = \frac{\sum_{i=1}^{n} Precision_i}{n} \tag{15}$$

$$Macro-averaged\ recall = \frac{\sum_{i=1}^{n} Recall_i}{n} \tag{16}$$

$$Micro-averaged\ precision = \frac{\sum_{i=1}^{n} TP_i}{\sum_{i=1}^{n}(TP_i+FP_i)} \tag{17}$$

$$Micro-averaged\ recall = \frac{\sum_{i=1}^{n} TP_i}{\sum_{i=1}^{n}(TP_i+FN_i)} \tag{18}$$

The macro average is calculated by simply taking the average obtained for each dataset, which gives an equal weight for every dataset and product. Whereas the micro average assigns each dataset a relative weight on the basis of the number of extracted or manually tagged aspects for the dataset.

## 4.5. Comparative study

In our experiments, first we evaluated Persian sentiment classification in two phases:
  Phase 1. *Without n-gram features and stemming*
  Phase 2. *With n-gram features and stemming*

Table 6 shows the F-score results for the two phases. From the results we can observe that using of n-gram features and stemming for sentiment classification has 4% and 0.3% improvements for negative and positive classes respectively.

**Table 2.** F-scores for phases 1 and 2, without and with n-gram features and stemming

| Phase | Class | F-score |
|---|---|---|
| 1 | Negative | 0.7480 |
|   | Positive | 0.8570 |
| 2 | Negative | 0.7880 |
|   | Positive | 0.8600 |

In this work we applied four different feature selection approaches, MI, DF, TFV and MMI with the Naive Bayes learning algorithm to the online Persian cellphone reviews. In the experiments, we found that using feature selection with learning algorithms can perform improvement to classifications of sentiment polarities of reviews.

Table 7 indicates Precision, Recall and F-score measures on two classes of Positive and Negative polarity with the feature selection approaches.

**Table 3.** Precision, Recall and F-score measures for the feature selection approaches with naive bayes classifier

| Approach | Class | Precision | Recall | F-score |
|---|---|---|---|---|
| **MI** | Negative | 0.4738 | 0.8356 | 0.6026 |
|  | Positive | 0.8130 | 0.4260 | 0.5538 |
| **DF** | Negative | 0.8148 | 0.7812 | 0.7962 |
|  | Positive | 0.8692 | 0.8898 | 0.8788 |
| **TFV** | Negative | 0.8226 | 0.7800 | 0.7996 |
|  | Positive | 0.8680 | 0.8956 | 0.8814 |
| **MMI** | Negative | 0.7842 | 0.8568 | 0.8172 |
| **(Proposed Approach)** | Positive | 0.9072 | 0.8526 | 0.8784 |

The results from Table 7 indicate that the TFV, DF and MMI have better performances than the traditional MI approach. In terms of F-score, MMI improves MI with 21.46% and 32.46% on Negative and Positive classes respectively, DF overcomes MI with 19.36% and 32.5% better performances for Negative and Positive review documents respectively and TFV improves MI with19.7% and 32.76% for Negative and Positive documents respectively. The reason of poor performance for MI is that of MI only uses the information between the corresponding feature and the corresponding class and does not utilize other information about other features and other classes. When we compare DF, TFV and MMI, we can find that the MMI beats both DF and TFV on F-scores of Negative review documents with 2.1% and 1.76% improvements respectively, but for the Positive review documents DF and TFV have 0.04% and 0.3% better performance than the MMI, respectively.

To assess the overall performance of techniques we adopt the macro and micro average, Figures 5, 6 and 7 show the macro and micro average for precision, recall and F-score respectively.

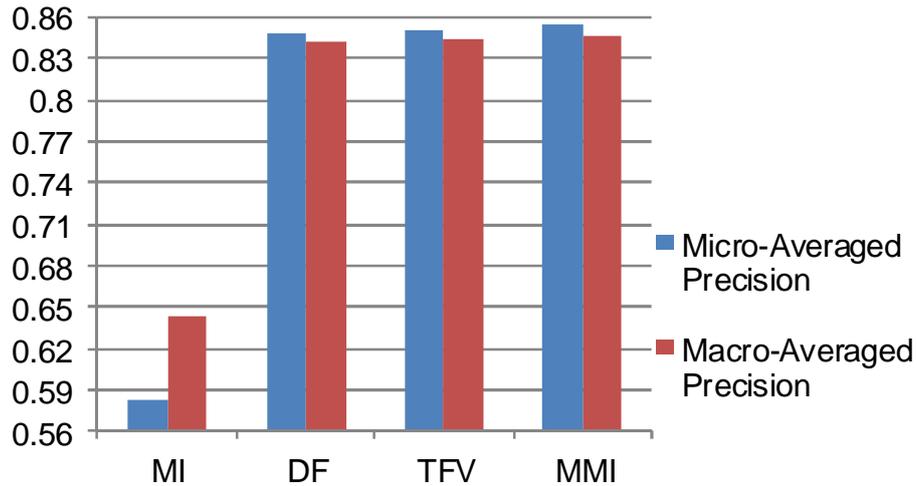

**Fig. 1.** Macro and micro average for precision on four approaches, MI, DF, TFV and MMI

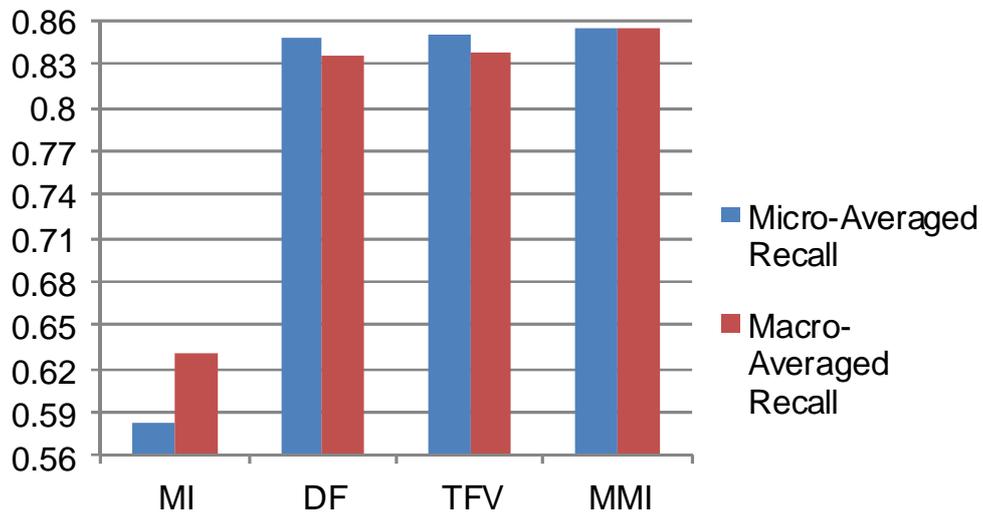

**Fig. 2.** Macro and micro average for recall on four approaches, MI, DF, TFV and MMI

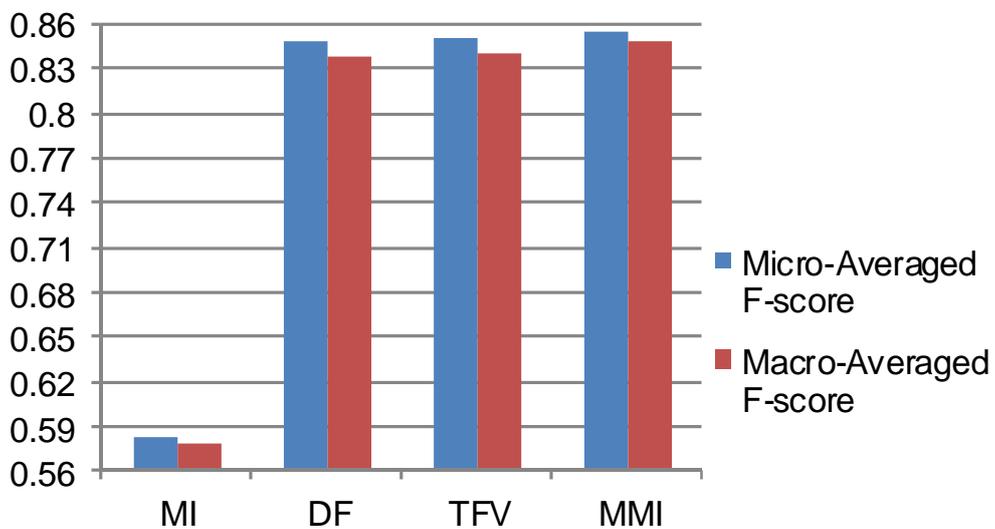

**Fig. 3.** Macro and micro average for F-score on four approaches, MI, DF, TFV and MMI

From these figures we can find that the MMI proposed approach has slightly better performance than the DF and TFV approaches and has significant improvements on MI method. The basic advantage of the MMI is using of whole information about a feature, positive and negative factors between features and classes. MMI in overall can reach to 85% of F-score classification.

It is worth noting that with a larger training corpus the feature selection approaches and the learning algorithm could get higher performance values. Additionally the proposed approach – MMI – is not only for Persian reviews and in addition can be applied to other domains or other classification problems.

Finally we can conclude that the proposed MMI algorithm is a promising alternative algorithm for the feature selection problem in text mining area of research. MMI considers all possible combinations of co-occurrences of a feature and a class label with computing positive and negative factors to calculate the score between the feature and class label. The proposed MMI approach can significantly improve previous standard methods while the approach is applicable for other domains of data mining and machine learning.

## 5. CONCLUSION AND FUTURE WORKS

This work presents a study on four feature selection approaches in sentiment analysis for Persian documents. In this paper we proposed a novel approach for feature selection, MMI, in sentiment classification problem. In addition we applied three other feature selection approaches, DF, MI and TFV with the Naive Bayes learning algorithm to the online Persian cellphone reviews. As the results show, using feature selection in sentiment analysis can improve the performance. The proposed MMI method that uses the positive and negative factors between features and classes improves significantly the performance compared to the other approaches. Based on the definition of MMI, it is a promising approach which can be utilized in other applications of data mining. In our future work we will focus more on sentiment analysis about Persian text. We will extend our model of classification to the word-level analysis and we will study approaches for opinion target identification and opinion word detection for Persian sentiment analysis. The main limitation of the study on Persian text is lack of data resource and open-source tools.

## REFERENCES

Ali, M. Z., Alkhatib K., Tashtoush Y., 2013, Cultural algorithms: emerging social structures for the solution of complex optimization problems, International Journal of Artificial Intelligence, **11**(A13), 20-42.

Abbasi, A., Chen, H., Salem, A., 2008, Sentiment analysis in multiple languages: feature selection for opinion classification in web forums, ACM Transactions on Information Systems (TOIS), **26**(3), 12.

Bagheri, A., Akbarzadeh - T, M. R., Saraee, M., 2008, Finding shortest path with learning algorithms, International Journal of Artificial Intelligence, **1**(A08), 86-95.

Bagheri, A., Saraee, M., de Jong, F., 2013a, Care more about customers: unsupervised domain-independent aspect detection for sentiment analysis of customer reviews, Knowledge-Based Systems, **52**, 201-213.

Bagheri, A., Saraee, M., de Jong, F., 2013b, An unsupervised aspect detection model for sentiment analysis of reviews, Proceedings of Natural Language Processing and Information Systems, Springer Berlin Heidelberg, 140-151.

Cui, H., Mittal, V., Datar, M., 2006, Comparative experiments on sentiment classification for online product reviews, Proceedings of National Conference on Artificial Intelligence, Menlo Park, Cambridge, London, **21**(2), 1265.

Duric, A., Song, F., 2012, Feature selection for sentiment analysis based on content and syntax models, Decision Support Systems, **53**(4), 704-711..

Farhoodi, M., Yari A., 2010, Applying machine learning algorithms for automatic Persian text classification, Proceedings of IEEE International Conference on Advanced Information Management and Service, 318-323.

Gamon, M. 2004, Sentiment classification on customer feedback data: noisy data, large feature vectors, and the role of linguistic analysis, Proceedings of 20th International Conference on Computational Linguistics, 841.

Ghani, R., Probst, K., Liu, Y., Krema, M., Fano, A. 2006, Text mining for product attribute extraction. SIGKDD Explorations, **8**(1), 41-48.

Hogenboom, A., Boon, F., Frasincar, F., 2012, A Statistical approach to star rating classification of sentiment. Management Intelligent Systems, 251-260.

Hu, M., Liu, B., 2004a, Mining opinion features in customer reviews, Proceedings of 19th National Conference on Artificial Intelligence AAAI, **4**, 755-760.

Hu, M., Liu, B., 2004b, Mining and summarizing customer reviews, Proceedings of ACM SIGKDD International Conference on Knowledge Discovery and Data Mining, 168-177.

Liu, B., Hu, M., Cheng, J., 2005, Opinion observer: analyzing and comparing opinions on the web, Proceedings of International Conference on World Wide Web, 342-351.

Liu, B., Zhang, L., 2012, A survey of opinion mining and sentiment analysis, Mining Text Data, 415-463.

Mitchell, T., 1997, Machine Learning, second edition, McGraw Hill.

Moraes, R., Valiati, J. F., Gavião Neto, W. P., 2012, Document-level sentiment classification: an empirical comparison between SVM and ANN, Expert Systems with Applications, **40**(2):621-633.

Pang, B., Lee, L., Vaithyanathan S., 2002, Thumbs up?: sentiment classification using machine learning techniques, Proceedings of the ACL-02 Conference on Empirical methods in Natural Language Processing, 10, 79-86.

Pei, Z., Shi, X., Marchese, M., Liang, Y., 2007, Text categorization method based on improved mutual information and characteristic weights evaluation algorithms, Proceedings of Fourth International Conference on Fuzzy Systems and Knowledge Discovery, Haikou, China, 87-91.

Popescu, A.M., Etzioni, O., 2007, Extracting product features and opinions from reviews, Proceedings of Conference on Empirical Methods in Natural language processing and text mining, Springer London, 9-28.

Precup, R.-E., David R.-C., Petriu E. M., Preitl S., Radac M.-B., 2012, Novel adaptive gravitational search algorithm for fuzzy controlled servo systems, IEEE Transactions on Industrial Informatics, **8**(4), 791-800.

Qiu, G., Liu, B., Bu, J., Chen, C., 2011, Opinion word expansion and target extraction through double propagation, Computational Linguistics, **37**(1), 9-27.

Rennie, J., 2004, *Improving multi-class text classification with Naive Bayes*, Massachusetts institute of technology, artificial intelligence laboratory, AI Technical Report.

Saraee, M., Bagheri, A., 2013, Feature selection methods in Persian sentiment analysis, Proceeding of Natural Language Processing and Information Systems, Springer Berlin Heidelberg, 303-308.

Shams, M., Shakery, A., Faili, H., 2012, A non-parametric LDA-based induction method for sentiment analysis, Proceedings of 16th IEEE CSI International Symposium on Artificial Intelligence and Signal Processing, 216-221.

Taghva, K., Beckley, R., Sadeh, M., 2005, A stemming algorithm for the Farsi language, Proceedings of IEEE International Conference on Information Technology: Coding and Computing, ITCC. vol. 1, 158-162.

Tang, B., Shepherd, M., Milios, E., Heywood, M. I., 2005, Comparing and combining dimension reduction techniques for efficient text clustering, Proceedings of International Workshop on Feature Selection for Data Mining.

Turney, P. D., Littman, M. L., 2002, *Unsupervised learning of semantic orientation from a hundred-billion-word corpus*, Technical Report EGB-1094, National Research Council Canada.

Yang, Y., Pedersen, J. O. 1997, A comparative study on feature selection in text categorization, Proceedings of 14th International Conference on Machine Learning ICML, 412–420.

Yang ,Y., 1999, An evaluation of statistical approaches to text categorization, Information retrieval, **1**(1), 69-90.

Yi, J., Nasukawa, T., Bunescu, R., Niblack, W., 2003, Sentiment analyzer: Extracting sentiments about a given topic using natural language processing techniques, Proceedings of the 3rd IEEE International Conference on Data Mining, 427–434.

Yussupova, N., Bogdanova, D., Boyko, M., 2012, Applying of sentiment analysis for texts in Russian based on machine learning approach, Proceedings of Second International Conference on Advances in Information Mining and Management, 8-14.

Zhu, J., Wang, H., Mao, J.T., 2010, Sentiment classification using genetic algorithm and Conditional Random Fields, Proceedings of 2nd IEEE International. Conference on Information Management and Engineering, 193-196.

Zhu, J., Wang, H., Zhu, M., Tsou, B.K., Ma, M., 2011, Aspect-based opinion polling from customer reviews, IEEE Transactions on Affective Computing, **2**(1), 37-49.